# PAC-learning bounded tree-width Graphical Models

**Mukund Narasimhan**[*]
Dept. of Electrical Engineering
University of Washington
Seattle, WA 98195

**Jeff Bilmes**[*]
Dept. of Electrical Engineering
University of Washington
Seattle, WA 98195

## Abstract

We show that the class of strongly connected graphical models with tree-width at most $k$ can be properly efficiently PAC-learnt with respect to the Kullback-Leibler Divergence. Previous approaches to this problem, such as those of Chow ([1]), and Hoffgen ([7]) have shown that this class is PAC-learnable by reducing it to a combinatorial optimization problem. However, for $k > 1$, this problem is NP-complete ([15]), and so unless P=NP, these approaches will take exponential amounts of time. Our approach differs significantly from these, in that it first attempts to find approximate conditional independencies by solving (polynomially many) submodular optimization problems, and then using a dynamic programming formulation to combine the approximate conditional independence information to derive a graphical model with underlying graph of the tree-width specified. This gives us an efficient (polynomial time in the number of random variables) PAC-learning algorithm which requires only polynomial number of samples of the true distribution, and only polynomial running time.

## 1 Introduction and Previous work

Let $(P, G)$ be a graphical model, where $G$ is a graph, and $P$ is a probability distribution over random variables corresponding to the vertices of $G$. If only the probabilty distribution $P$ is specified, then in general, there is no unique graph $G$ over which $P$ factorizes. In fact, all probability distributions factorize over the complete graph. The goal therefore, is to find a graph $G$ with as few edges as possible over which the given distribution factorizes. As this problem is NP-complete ([11]), we look at the following variant of this problem. Given a class of graphs $\mathcal{G}$, where it is known that the distribution $P$ factorizes over at least one graph $G \in \mathcal{G}$, find a graph in $\mathcal{G}$ over which $P$ factorizes. The complexity of this problem depends on the class $\mathcal{G}$, and much research has been directed at finding classes $\mathcal{G}$ which can be learnt in polynomial time. There is a plethora of negative results ([8, 11, 14, 16]) showing that learning various classes of graphs, including paths and polytrees, is NP-complete. So far the only positive result in this area (to the best of our knowledge) are the results of ([1, 7]) in which it is shown that the class of trees can be properly efficiently learnt. While trees are an important class of distributions, they can only represent very sparse dependencies. There is much interest in learning other model families in which richer dependence structures are possible.

Prior attempts at learning the model family rely on reducing the problem to a combinatorial optimization problem. When the combinatorial optimization problem is easy (such as finding minimum/maximum spanning tree), then the learning problem becomes easy. The reverse reduction (reducing an arbitrary instance of a NP-complete problem to a learning problem) is used to demonstrate hardness of learning problem. We use a different approach, and show the learnability of a subclass of bounded tree-width networks by first discovering the conditional independencies in the model by solving a polynomial number of submodular optimization problems. Each one of these problems can be solved in polynomial time using an algorithm developed by Queyranne ([10]). We then use a dynamic programming algorithm to patch these dependencies together to create a graph of the tree-width specified. While we restrict ourselves to showing the learnability of (partial) $k$-trees, this approach could conceivably be used for other subclasses of graphical models as well.

---

[*]This work was supported by NSF grant IIS-0093430 and an Intel Corporation Grant

## 2　Preliminaries and Notation

A graph $G = (V, E)$ is said to be *chordal* if every cycle of length at least 4 has a chord. A *tree-decomposition*([6]) of a graph $G = (V, E)$ is a pair $(T = (I, F), \{V_i\}_{i \in I})$, where $T$ is a tree, and $V(G) = \cup_{i \in I} V_i$ such that if $uv$ is an edge in $E(G)$, then there is a $i \in I$ such that $u, v \in V_i$, and if $i, j, k \in I$ are such that $j$ is on the path from $i$ to $k$ in $T$, then $V_i \cap V_k \subseteq V_j$. The *tree-width* of this tree-decomposition is $\max_{i \in I} |V_i| - 1$. A *partial $k$-tree* is a chordal graph which has a tree-decomposition with width $k$. For any edge $e = \{i, j\} \in F$, we let $V_e = V_{ij} = V_i \cap V_j$. Note that $V_e$ is a separator in $G$ for every $e \in F$, and in fact, these are the unique minimal separators of $G$.

Let $P$ be a probability distribution over random variables $\{X_v\}_{v \in V}$. For any $A \subseteq V$, we let $P_A$ be the (marginalized) probability distribution of the random vector $(X_a)_{a \in A}$. We say that the probability distribution *factorizes* according to the chordal graph $G = (V, E)$ if for every minimal separator $S \subseteq V$ which separates the graph into components $A$ and $B$, the marginal distributions $P_{A \cup S}$ and $P_{B \cup S}$ factorize with respect to $G[A \cup S]$ and $G[B \cup S]$ respectively. Let us denote by $H_P(A)$ the binary entropy $H(\{X_v\}_{v \in A})$ with respect to the probability distribution $P$, and by $I_P(A; B|S)$ the mutual information $I(\{X_v\}_{v \in A}; \{X_v\}_{v \in B} | \{X_v\}_{v \in S})$ with respect to $P$. If $P$ factorizes over $G$, and $S$ is a minimal separator of $G$ such that $G[V \setminus S]$ has connected components $\{C_1, C_2, \ldots, C_m\}$, then $I_P(V(C_i); V(C_j)|S) = 0$ for all $i \neq j$. $S$ is called an $\alpha$-strong separator if we cannot partition the vertices in $C_i$ into $C_i = C_{i1} \cup C_{i2}$ satisfying $I_P(C_{i1}; C_{i2}|S) \leq \alpha$. We say that the graphical model $(P, G)$ is $\alpha$-strongly connected if every minimal separator $S$ of the chordal graph $G$ is an $\alpha$-strong separator. We say that $(P, G)$ is strongly connected if there is some $\alpha > 0$ such that every minimal separator is an $\alpha$-strong separator.

A class of distributions $\mathcal{T}$ is *properly efficiently learnable* (with respect to the KL-divergence criteria) if there exists a probabilistic algorithm which for every $P \in \mathcal{T}$, and for every $\epsilon, \delta > 0$, finds a $\tilde{P} \in \mathcal{T}$ such that $D(P \| \tilde{P}) < \epsilon$ with probability at least $1 - \delta$ in time polynomial in $n$, $\frac{1}{\epsilon}$ and $\frac{1}{\delta}$. The algorithm may sample from the (true) distribution $P$, but at most a polynomial number of times. We will assume the existence of an algorithm that can estimate the entropies of an arbitrary subset of the random variables with precision $\epsilon$ and confidence $1 - \delta$ using $f(n, \epsilon, \delta)$ samples, where $f(n, \epsilon, \delta)$ is polynomial in $n$, $\frac{1}{\epsilon}$ and $\frac{1}{\delta}$. The exact algorithm might depend on the nature of the distribution (i.e., depending on whether the random variables are discrete, continuous, Gaussian etc.). An application of Hoeffding's inequality shows that this is possible for discrete distributions (alternatively see [7]). It is also possible to do this for continuous/mixed distributions by discretizing, though much more efficient algorithms exist for distributions such as Gaussians. It is clear that if each query can be computed in polynomial time, with at most polynomial number of samples, then any algorithm which issues at most a polynomial number of such queries, and runs in polynomial time other than these queries is in fact a polynomial time algorithm.

## 3　Submodularity and Partitions

A set function $f : 2^V \to \mathbb{R}^+$ is called *submodular* if for any $A, B \subseteq V$, $f(A) + f(B) \geq f(A \cup B) + f(A \cap B)$. $f$ is said to be *symmetric* if $f(A) = f(V \setminus A)$. Submodular functions may be thought of as the discrete analog of convex functions, and have several similar properties. For example, there are polynomial time algorithms to minimize submodular functions (see [3]). Our interest in submodular functions stems from the fact that the mutual information function is a submodular function.

**Lemma 1.** *Let $P$ be any probability distribution, and $S \subseteq V$ be a fixed set. Then $H_P(\cdot|S) : 2^V \to \mathbb{R}$ is a submodular function.*

*Proof.* $0 \leq I_P(A;B|S) = H_P(A|S) + H_P(B|S) - H_P(A \cup B|S) - H_P(A \cap B|S)$. □

**Proposition 2.** *Let $P$ be an arbitrary distribution on the random variables $\{X_v\}_{v \in V}$. Let $F : 2^V \to \mathbb{R}^+$ be given by $F(A) = I_P(A; V \setminus A)$. Then $F$ is symmetric and submodular.*

*Proof.* We can write $F(A) = H_P(A) + H_P(V \setminus A) - H_P(V)$. Therefore, $F(A) + F(B) = [H_P(A) + H_P(B)] + [H_P(V \setminus A) + H_P(V \setminus B)] - 2H_P(V)$. By the submodularity of $H_P(\cdot)$, $H_P(A) + H_P(B) \geq H_P(A \cap B) + H_P(A \cup B)$. Similarly, $H_P(V \setminus A) + H_P(V \setminus B) \geq H_P(V \setminus (A \cup B)) + H_P(V \setminus (A \cap B))$. The result now follows by noting that $F(A \cap B) = H_P(A \cap B) + H_P(V \setminus (A \cap B)) - H_P(V)$ and $F(A \cup B) = H_P(A \cup B) + H_P(V \setminus (A \cup B)) - H_P(V)$. Therefore $F(\cdot)$ is submodular and symmetric. □

We will be investigating conditional independence relationships between random variables. If a set of random variables $\{X_a\}_{a \in A}$ is conditionally independent of the set of random variables $\{X_b\}_{b \in B}$, given the set $\{X_s\}_{s \in S}$ according to the probability distribution $P$, then $I_P(A;B|S) = 0$. Note that in this definition, we have not required that $A$ and $B$ be disjoint (or $A \cap B = \phi$). However since $I_P(A \cup S; B \cup S|S) = I_P(A; B|S)$ it will be convenient for our application to let $A \cap B = S$. Let $V_{\overline{S}} = V \setminus S$. The following corollary is immediate.

**Corollary 3.** *For any probability distribution $P$, the function $F_{P,S} : 2^{V_{\overline{S}}} \to \mathbb{R}^+$, given by $F_{P,S}(A) = I_P(A; V_{\overline{S}} \setminus A|S)$, is symmetric and submodular.*

One reason for the importance of submodular functions is the existence of efficient algorithms to minimize submodular set functions (see [2]). If $f : 2^V \to \mathbb{R}^+$ is both symmetric and submodular, then there exist combinatorial algorithms to compute this minimum. For example, Queyranne's Algorithm, described in [10], will return a proper subset $A \subseteq V$ such that $A \in \arg\min_{B \in 2^V \setminus \{V, \phi\}} f(B)$. Let us denote this algorithm by QA. QA takes as input a $f$-value oracle, and runs in time $O(|V_{\overline{S}}|^3)$, using at most $|V_{\overline{S}}|^3$ oracle calls. In particular, since $F_{P,S} : 2^{V_{\overline{S}}} \to \mathbb{R}^+$ is symmetric and submodular, we can (quickly) find a set $A$ such that $F_{P,S}(A)$ is minimized. If $S \subseteq V$, then we say that $S$ is a $\epsilon$-separator of $V$ if there is a partition $A \cup B$ of $V_{\overline{S}}$ such that $I_P(A;B|S) \leq \epsilon$. In this case, we call $\{A, B\}$ an $\epsilon$-partition for $(V_{\overline{S}}, S, P)$. Clearly, we can use QA, and a $F_{P,S}$-value oracle, to determine if there are any $\epsilon$-partitions for $(V_{\overline{S}}, S, P)$.

When we are trying to estimate or learn a probability distribution $P$ by sampling, we do not have a $F_{P,S}$-value oracle. However, sampling will let us estimate $F_{P,S}$. So, suppose that we had a $\widetilde{F_{P,S}}$-value oracle, which satisfies $\left|\widetilde{F_{P,S}}(A) - F_{P,S}(A)\right| \leq \epsilon_1$ for every $A \subseteq V_{\overline{S}}$. One problem is that $\widetilde{F_{P,S}}(A)$ need not be submodular, and hence QA need not return the minimum value of $\widetilde{F_{P,S}}(A)$. However, we do have the following result.

**Lemma 4.** *Suppose that $F_{P,S} : 2^{V_{\overline{S}}} \to \mathbb{R}^+$ is a symmetric submodular function, and $\widetilde{F_{P,S}} : 2^{V_{\overline{S}}} \to \mathbb{R}^+$ is another function that is not necessarily submodular, but satisfies $\left|\widetilde{F_{P,S}}(A) - F_{P,S}(A)\right| \leq \epsilon_1$ for every $A \subseteq V_{\overline{S}}$. Then QA will return a non-empty proper subset $\tilde{A} \subset V_{\overline{S}}$ such that for every non-empty proper subset $A \subset V_{\overline{S}}$, $\widetilde{F_{P,S}}(\tilde{A}) - \widetilde{F_{P,S}}(A) \leq |V_{\overline{S}}| \cdot \epsilon_1 \leq |V| \cdot \epsilon_1$*

The proof of this lemma can be found in the appendix. So, $\tilde{A}$ is an approximate minimizer for $\widetilde{F_{P,S}}(\cdot)$. In fact it is also an approximate minimizer for $F_{P,S}(\cdot)$.

**Corollary 5.** *Suppose that $A$ minimizes $F_{P,S}(\cdot)$, and $\tilde{A}$ is the set returned by QA. Then $F_{P,S}(\tilde{A}) - F_{P,S}(A) \leq (|V| + 2)\epsilon_1$.*

*Proof.*

$$\begin{aligned}
0 &\leq F_{P,S}(\tilde{A}) - F_{P,S}(A) \\
&\leq (\widetilde{F_{P,S}}(\tilde{A}) + \epsilon_1) - (\widetilde{F_{P,S}}(A) - \epsilon_1) \\
&= 2\epsilon_1 + \widetilde{F_{P,S}}(\tilde{A}) - \widetilde{F_{P,S}}(A) \\
&\leq 2\epsilon_1 + |V| \cdot \epsilon_1 \\
&= (|V| + 2)\epsilon_1
\end{aligned}$$
□

Therefore, if we run QA using the $\widetilde{F_{Q,S}}$-value oracle, we will get a set $\tilde{A}$ that is $(|V|+2)\epsilon_1$-close to optimal for $F_{P,S}$. In particular, for all proper subsets $B \subseteq V_{\overline{S}}$, $F_{P,S}(B) \geq F_{P,S}(\tilde{A}) - (|V|+2)\epsilon_1$. A sufficient condition for this to hold is $\left|F_{P,S}(A) - \widetilde{F_{P,S}}(A)\right| \leq \epsilon_1$ for every query issued by the run of QA. If each oracle query $\widetilde{F_{P,S}}(A)$ can be computed in polynomial time, and using only polynomially many samples, with precision $\epsilon_1$ and confidence at least $1 - \delta_1$, then $\left|\widetilde{F_{P,S}}(A) - F_{P,S}(A)\right| \leq \epsilon_1$ for all the $A$'s queried by the run of QA with confidence at least $1 - \left|V_{\overline{S}}\right|^3 \delta_1$ by the union bound[1] We summarize this discussion in the proposition below

**Proposition 6.** *Suppose that $\widetilde{F_{P,S}}$ represents an approximation to $F_{P,S}$ which satisfies $\left|\widetilde{F_{P,S}}(A) - F_{P,S}(A)\right| \leq \epsilon_1$ with confidence at least $1 - \delta_1$ for any (particular) $A \subseteq V_{\overline{S}}$. Then running QA with a $\widetilde{F_{P,S}}$-value oracle results in a solution $\tilde{A} \subseteq V_{\overline{S}}$ which is within $(|V|+2)\epsilon_1$ close to optimal (for $F_{P,S}$) with confidence at least $1 - \left|V_{\overline{S}}\right|^3 \delta_1$. In particular, if $\widetilde{F_{P,S}}(\tilde{A}) \geq (|V|+2)\epsilon_1 + \epsilon_2$, then there is no $\epsilon_2$-partition for $(V_{\overline{S}}, S, P)$ (with confidence at least $1 - \left|V_{\overline{S}}\right|^3 \delta_1$).*

The function $F_{P,S}(A) = I_P(A; V_{\overline{S}} \setminus A | S)$, represents the mutual information between the random variables indexed by $A$ and the remaining random variables once $S$ is given. We will call $I_{\tilde{P}}(\cdot | S)$ or $\widetilde{F_{P,S}}$ the sample or the approximate mutual information, corresponding to sample distribution $\tilde{P}$. Suppose that $A \cup B = V_{\overline{S}}$. If $I_{\tilde{P}}(A; B | S) \leq \epsilon$, we will say that $\{A, B\}$ is an $\epsilon$-partition for $(V_{\overline{S}}, S, \tilde{P})$. Now, a graphical model representing $P$ can be thought of as representing conditional independencies, and correspondingly, our goal is to learn these conditional independencies. If $A \perp\!\!\!\perp B | S$, then $I_P(A; B | S) = 0$. This of course means that $\{A, B\}$ is an 0-partition for $(V_{\overline{S}}, S, P)$. Hence it is a $\epsilon_1$-partition for $(V_{\overline{S}}, S, \tilde{P})$ with high confidence. We should note however, that the converse is not necessarily true. If $\{A, B\}$ is a $\epsilon_1$-partition for $(V_{\overline{S}}, S, \tilde{P})$, then it need not be a 0-partition for $(V_{\overline{S}}, S, P)$. However, it is a $(2\epsilon_1)$- partition for $(V_{\overline{S}}, S, P)$ with high confidence, and so for sufficiently small $\epsilon_1$, $A$ and $B$ are "almost" independent given $S$. In fact, as we will show in a later section, we can always find almost independent partitions so that the total KL divergence between the resulting distribution, and the original distribution remains small.

---
[1] If each $A_i$ occurs with confidence $1 - p_i$, then $A_i$ does not occur with probability $p_i$. Therefore, the probability that at least one of the $A_i$ does not occur is at most $\sum p_i$. Hence all the $A_i$ occur with probability at least $1 - \sum p_i$.

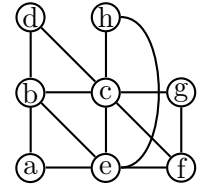

Figure 1: $S = \{c, e\}, \pi_S = \{\{g, f\}, \{a, b, d\}, \{h\}\}$

If $G$ is a graphical model representing the probability distribution $P$, and $A \perp\!\!\!\perp B | S$, then $G[V \setminus S]$ is a disconnected graph, and $A$ and $B$ are the union of components of $G[V \setminus S]$. For example, in the graph in Figure 1, if $S = \{c, e\}$, then the removal of $S$ disconnects the graph into the parts $\{a, b, d\}$ and $\{g, f, h\}$. Note that $\{g, f, h\}$ is not connected in the residual graph, but is the union of $\{g, f\}$ and $\{h\}$. It helps to consider the components themselves, rather than the union of components, and so, we extend our notion of almost conditional independence to partitions of $V_{\overline{S}}$ of the form $\pi = \{A_1, A_2, \ldots, A_m\}$, where $m \geq 2$. $\pi$ is a partition of $V_{\overline{S}}$ and so $V_{\overline{S}}$ is the disjoint union $\cup_{i=1}^m A_i$. We call $\pi$ an $\epsilon$-partition for $(V_{\overline{S}}, S, P)$ if $I_P(A_i; A_j | S) \leq \epsilon$ for all $1 \leq i \neq j \leq m$.

**Lemma 7.** *If $\pi$ is a $\epsilon_2$-partition for $(V_{\overline{S}}, S, \tilde{P})$, and $A \subseteq V_{\overline{S}}$, then $\rho = \{A_i \cap A : A_i \in \pi\}$ is a $\epsilon_2$-partition for $(A, S, \tilde{P})$. Further, if $A \in \pi$, and $\rho$ is a $\epsilon_3$-partition for $(A, S, \tilde{P})$, then $\psi = (\pi \setminus \{A\}) \cup \rho$ is a $\max(\epsilon_2, \epsilon_3)$-partition for $(V_{\overline{S}}, S, \tilde{P})$.*

*Proof.* The first remark follows because $I_P(A_i \cap A; A_j \cap A | S) \leq I_P(A_i; A_j | S)$. If $A \in \pi$, then it can be checked that $\psi$ is a partition for $V_{\overline{S}}$. Let $\epsilon = \max(\epsilon_2, \epsilon_3)$. If $C, D \in \psi$, we consider 3 cases. First, if $C, D \in \pi$, then $I_{\tilde{P}}(C; D | S) \leq \epsilon_2 \leq \epsilon$. If $C, D \in \rho$, then $I_{\tilde{P}}(C; D | S) \leq \epsilon_3 \leq \epsilon$. If neither of these cases occur, we may assume (by symmetry) that $C \in \psi$ and $D \in \pi$. Then $I_{\tilde{P}}(C; D | S) \leq I_{\tilde{P}}(C; A | S) \leq \epsilon_2 \leq \epsilon$. Therefore $\psi$ is an $\epsilon$-partition for $(V_{\overline{S}}, S, \tilde{P})$. □

Lemma 7 suggests a possible strategy for finding the almost independent components, for a given fixed separator $S$. Start with the trivial partition $\pi = \{V_{\overline{S}}\}$ of $V_{\overline{S}}$, and iteratively attempt to partition the partitions, till the information loss is too large. This is shown in Algorithm 1. For any

---

$\pi_S^0 \leftarrow \{V_{\overline{S}}\}$; $i \leftarrow 0$;
**while** $\exists X_i \in \pi_S^i$ such that $\{A_i, B_i\}$ is an $\epsilon$-partition of $(X_i, S, \tilde{P})$ **do**
$\quad \pi_S^{i+1} \leftarrow (\pi_S^i \setminus \{X_i\}) \cup \{A_i, B_i\}$; $i \leftarrow i + 1$;
**end**
$\pi_S(\epsilon) \leftarrow \pi_S^i$

**Algorithm 1:** Algorithm to compute $\epsilon$-partition $\pi_S(\epsilon)$ (for any given $\epsilon$ and $S$)

---

$S \subseteq V$ and maximum allowable information loss $\epsilon > 0$, the algorithm shown computes a partition $\pi_S(\epsilon)$ of $V_{\overline{S}}$. We say that a partition $\psi$ is a refinement of $\pi$ if each element of $\pi$ can be written as the union of elements of $\psi$. Therefore, in Algorithm 1, the partition $\pi_S^i$ refines $\pi_S^j$ for all $j \leq i$. We then have the following result.

**Lemma 8.** *The partition $\pi_S(\cdot)$ produced by the above algorithm satisfies the following properties with confidence at least $1 - O(|V|^5)\delta_1$:*

1. *If $\{A, B\}$ is any $\epsilon_2$-partition of $(V_{\overline{S}}, S, P)$, $\{A, B\}$ is refined by $\pi_S(\epsilon_2 + (|V| + 2)\epsilon_1)$.*

2. *If $C, D \in \pi_S(\epsilon_2 + (|V| + 2)\epsilon_1)$, and $C \neq D$, then $I_P(C; D|S) \leq \epsilon_2 + (|V| + 3)\epsilon_1$.*

3. *If $\{C, D\}$ is any partition (of $V_{\overline{S}}$) that is refined by $\pi_S(\epsilon_2 + 2\epsilon_1)$, then $I_P(C; D|S) \leq |V|^2 (\epsilon_2 + (|V| + 3)\epsilon_1) \leq |V|^3 (\epsilon_2 + 3\epsilon_1)$.*

*Proof.* We note that the loop in the algorithm shown above iterates at most $|V_{\overline{S}}|$ times. During each iteration QA is called at most $|V_{\overline{S}}|$ times, and hence there are at most $|V_{\overline{S}}|^5$ queries of the form $\widetilde{F_{P,S}}(A)$ made by the algorithm shown. So, $\left|\widetilde{F_{P,S}}(A) - F_{P,S}(A)\right| \leq \epsilon_1$ for all queries $\widetilde{F_{P,S}}(A)$ by QA, with confidence at least $1 - (|V|^5)\delta_1$ by the union bound. In particular, if $\{A, B\}$ is an $\epsilon_2$-partition of $(V_{\overline{S}}, S, P)$, then with confidence at least $1 - |V|^5 \delta_1$, it has to be a $(\epsilon_2 + (|V| + 2)\epsilon_1)$-partition of $(V_{\overline{S}}, S, \tilde{P})$ by Lemma 7. Therefore, if $\pi_S(\epsilon_2 + (|V| + 2)\epsilon_1)$ is not a refinement of $\{A, B\}$, then the algorithm should not have stopped where it did, a contradiction. The second assertion follows by induction and Lemma 7. The final assertion now follows by noting that if $\Lambda, \Gamma$ are disjoint, then $I_P(\cup_{\alpha \in \Lambda} Y_\alpha; \cup_{\beta \in \Gamma} Y_\beta | S) \leq \sum_{\alpha \in \Lambda, \beta \in \Gamma} I_P(Y_\alpha; Y_\beta | S) \leq |V|^2 (\epsilon_2 + (|V| + 3)\epsilon_1)$. □

Let $\mathcal{P}(\epsilon_1, \epsilon_2)$ be the set of partitions $\{\pi_S(\epsilon_2 + (|V| + 2)\epsilon_1) : S \subseteq V, |S| \leq k\}$. There are at most $\binom{|V|}{\leq k} \in O(|V|^k)$ sets of size at most $k$. Then, we have the following result

**Corollary 9.** *We can generate the set of partitions $\mathcal{P}(\epsilon_1, \epsilon_2)$ in time $O\left(|V|^{k+5}\right)$, making at most $O\left(|V|^{k+5}\right)$ oracle calls. The set $\mathcal{P}(\epsilon_1, \epsilon_2)$ satisfies the following with confidence at least $1 - |V|^{k+5} \delta_1$.*

1. *If $\{A, B\}$ is any $\epsilon_2$-partition of $(V, S, P)$, then $\pi_S(\epsilon_2 + (|V| + 2)\epsilon_1) \in \mathcal{P}(\epsilon_1, \epsilon_2)$ refines $\{A, B\}$.*

2. *If $\{A, B\}$ is any partition that $\pi_S(\epsilon_2 + (|V| + 2)\epsilon_1) \in \mathcal{P}(\epsilon_1, \epsilon_2)$ refines, then $\{A, B\}$ is an $|V|^3 (\epsilon_2 + 3\epsilon_1)$-partition.*

## 4 Partitions and Tree-Decompositions

If $G = (V, E)$ is a chordal graph of tree-width $k$, then $G$ has a tree-decomposition $(T = (I, F), \{V_i\}_{i \in I})$ of width $k$. Conversely, given a tree-decomposition $(T = (I, F), \{V_i\}_{i \in I})$ of width $k$, we get a chordal graph $G = (V, E)$ of tree-width $k$. For any edge $V_i V_j \in F$, let $V_{ij} = V_i \cap V_j$. Then it follows from a result in [5] that the minimal separators of $G$ are precisely the sets of the form $V_{ij}$. Since $T$ is a tree, the removal of any edge $V_i V_j$ breaks the graph up into two trees. We will let $V_{ij}^A$ and $V_{ij}^B$ be the vertices of $G$ corresponding to the two trees except for $V_{ij}$. If $P$ factorizes over $G$, then $\left\{V_{ij}^A, V_{ij}^V\right\}$ is a 0-partition for $(V_{\overline{V_{ij}}}, V_{ij}, P)$. In fact, we have ([9])

$$P\left(\{X_v\}_{v \in V}\right) = \frac{\prod_{i \in I} P\left(\{X_v\}_{v \in V_i}\right)}{\prod_{V_i V_j \in F} P\left(\{X_v\}_{v \in V_{ij}}\right)}$$

If $P$ satisfies the above condition, we say that $P$ factorizes over the tree-decomposition ($T = (I, F), \{V_i\}_{i \in I}$). So, finding a chordal graph $G$ of width $k$ such that $P(\cdot)$ factorizes over $G$ is equivalent to finding a tree-decomposition over which $P(\cdot)$ factorizes. Now, we can ask a related question : Given a tree-decomposition ($T = (I, F), \{V_i\}_{i \in I}$), find a probability distribution $\tilde{P}(\cdot)$ which factorizes over this tree-decomposition such that $D(P\|\tilde{P})$ is minimized. When $P(\cdot)$ factorizes over $T$, then $P(\cdot)$ is the unique optimal solution. More generally, we have the following result.

**Lemma 10.** *Let $P$ be a probability distribution, and $(T = (I, F), \{V_i\}_{i \in I})$ a tree-decomposition. Then the unique minimizer of $D(P\|P_1)$ subject to the constraint that $P_1$ factorizes over $T$ is given by*

$$P_1\left(\{X_v\}_{v \in V}\right) = \frac{\prod_{i \in I} P\left(\{X_v\}_{v \in V_i}\right)}{\prod_{V_i V_j \in F} P\left(\{X_v\}_{v \in V_{ij}}\right)}$$

*Further,*

$$D(P\|P_1) = \sum_{V_i V_j \in F} I_P(V_{ij}^A; V_{ij}^B | V_{ij})$$

*Proof.* See [17]. □

For any set $S \subseteq V$ with $|S| \leq k$, let $\pi_S \triangleq \pi_S(\epsilon_2 + (|V| + 2)\epsilon_1) \in \mathcal{P}(\epsilon_1, \epsilon_2)$ be the partition corresponding to $S$. We say that the tree-decomposition ($T = (I, F), \{V_i\}_{i \in I}$) is compatible with $\mathcal{P}(\epsilon_1, \epsilon_2)$ if $\pi_{V_{ij}}$ refines $\left\{V_{ij}^A, V_{ij}^B\right\}$. We then have the following result.

**Lemma 11.** *Suppose that $(T = (I, F), \{V_i\}_{i \in I})$ is a tree-decomposition that is compatible with $\mathcal{P}(\epsilon_1, \epsilon_2)$. Let $P_1$ be the optimal distribution as in Lemma 10. Then $D(P\|P_1) \leq |V|^4 (\epsilon_2 + 3\epsilon_1)$.*

*Proof.* Let $V_i V_j$ be any edge in $T$. Then $\pi_{V_{ij}}$ refines $\left\{V_{ij}^A, V_{ij}^B\right\}$, and so by Lemma 9, we have $I(V_{ij}^A; V_{ij}^B | V_{ij}) \leq |V|^3 (\epsilon_2 + 3\epsilon + 1)$. Therefore, by Lemma 10, $D(P\|\tilde{P}) \leq (|I| - 1) \cdot |V|^3 (\epsilon_2 + 3\epsilon_1) \leq |V|^4 (\epsilon_2 + 3\epsilon_1)$. □

Pick $\alpha > 0$ so that $P$ is $\alpha$-strongly connected. Given arbitrary $\epsilon, \delta > 0$, let $\delta_1 = \frac{\delta}{|V|^5}$, and $\epsilon_1 =$ $\epsilon_2 < \frac{\min(\epsilon, \alpha)}{4|V|^4}$. Then $|V|^4 (\epsilon_2 + 3\epsilon_1) < \min(\epsilon, \alpha)$. In particular, if $(T = (I, F), \{V_i\}_{i \in I})$ is any tree-decomposition for $P$, then for every separator $V_{ij}$, $\left\{V_{ij}^A, V_{ij}^V\right\}$ is refined by $\pi_{V_{ij}}$. Hence the tree-decomposition is compatible with $\mathcal{P}(\epsilon_1, \epsilon_2)$. Therefore, if there is a tree-decomposition over which $P$ factorizes, then it is compatible with $\mathcal{P}(\epsilon_1, \epsilon_2)$. Lemma 11 guarantees that if we restrict our search to tree-decompositions that are compatible with $\mathcal{P}(\epsilon_1, \epsilon_2)$, we are guaranteed to find a tree-decomposition for which the optimal probability distribution $P_1$ satisfies $D(P\|P_1) \leq |V|^4 (\epsilon_2 + 3\epsilon_1) < \epsilon$, since it is known that there is at least one tree-decomposition compatible with $\mathcal{P}(\epsilon_1, \epsilon_2)$. The obvious approach to finding such compatible tree-decompositions using dynamic programming by examining partitions in order of increasing size. Let $\mathcal{S}$ be the set of all pairs of the form $(S, A)$ where $A \in \pi_S \in \mathcal{P}(\epsilon_1, \epsilon_2)$ and $|S| \leq k$. Then $A$ is a element of the partition induced by (the separator) $S$, and we will define the size of $(S, A)$ to be $|(S, A)| = |S \cup A|$. A dynamic programming algorithm which works by examining parititons in order of increasing size is shown as Algorithm 2. This is a simple variant of the algorithm described in [4].

---

$\mathcal{S} \leftarrow \{(S, A) : |S| \leq k, A \in \pi_S \in \mathcal{P}(\epsilon_1, \epsilon_2)\}$;
$\mathcal{M} \leftarrow \{(S, A) \in \mathcal{S} : |S \cup A| \leq k + 1\}$;
**for** $(S, A) \in \mathcal{S}$ *in order of increasing* $|S \cup A|$ **do**
　**for** $v \in A \cup S$ **do**
　　$\mathcal{R} \leftarrow \left\{(L, B) \in \mathcal{M} : \begin{array}{l} L \subseteq S \cup \{v\} \\ B \cup L \subseteq A \cup S \end{array}\right\}$;
　　**if** $\cup_{(L,B) \in \mathcal{R}}(B \cup L) = A \cup S$ **then**
　　　$\mathcal{M} \leftarrow \mathcal{M} \cup \{(S, A)\}$;
　　**end**
　**end**
　**if** $\exists (S, \pi_S)$ *s.t.* $(S, A) \in \mathcal{M}\, \forall A \in \pi_S$ **then**
　　Exit("Tree-decomposition exists");
　**else**
　　Exit("No Tree-decomposition exits");
　**end**
**end**

**Algorithm 2:** Finding a Tree-Decomposition compatible with $\mathcal{P}(\epsilon_1, \epsilon_2)$.

**Theorem 12.** *For any $\epsilon, \delta > 0$, pick $\epsilon_1, \epsilon_2$ as above. Then computing $\mathcal{P}(\epsilon_1, \epsilon_2)$ using Algo-* *rithm 1, and then running Algorithm 2 will produce a tree decomposition, and hence a graphical model $(G, P_1)$ such that $D(P\|P_1) \leq \epsilon$ with confidence at least $\delta$ as long as $P$ factorizes over some graph of treewidth at most $k$.*

## 5  Conclusions

We have shown that by identifying approximate conditional independencies in a graphical model, we can identify the potential separators in the underlying graph. This yields a polynomial time algorithm to properly efficiently PAC-learn strongly connected graphical models of bounded tree-width. Interesting avenues for future research include investigating if this technique can be used for robust learning of larger classes of graphical models.

## A  Proof of Lemma 4

An ordered pair $(t, u)$ called a pendent pair for $(V, f)$ if for every $U \subseteq V$ with $u \in U$ and $t \notin U$, we have $f[u] \leq f[U]$. Algorithm 3 is a subroutine used in Queyranne's algorithm. When $f$ is symmetric and submodular, the subroutine returns a pendant pair $(t, u)$. We consider the case when $f$ is approximately symmetric and submodular. By this we mean that there is an $\epsilon > 0$ such that $f(A) + f(B) \geq f(A \cup B) + f(A \cap B) - \epsilon$, and $|f(A) - f(V \setminus A)| \leq \epsilon$ for every $A, B \subseteq V$. We then have the following result, which is equivalent to Theorem 2 in [10], which is proven using a variant of their argument.

---

$v_1 \leftarrow$ arbitrary $x$; $W_1 \leftarrow \{v_1\}$
**for** $i = 1 : n - 1$ **do**
$\quad \forall u \in V \setminus W_i$ **do** $\mathsf{key}[u] \leftarrow f[W_i \cup u] - f[u]$;
$\quad v_{i+1} \leftarrow \arg\min_{u \in V \setminus W_i} \mathsf{key}[u]$
$\quad W_{i+1} \leftarrow W_i \cup v_{i+1}$
**end**
**return**$(v_{n-1}, v_n)$

---

**Algorithm 3:** Compute a pendent pair $(t, v) = (v_{n-1}, v_n)$.

**Lemma 13.** *Suppose that $f : 2^V \to \mathbb{R}$ is approximately symmetric and submodular. Then for every $1 \leq i \leq n-1$, for each $u \notin W_i$ and $S \subseteq W_{i-1}$, we have $f[W_i] + f[u] \leq f[W_i \setminus S] + f[X \cup u] + (i-1)\epsilon$*

*Proof.* We show this by induction. For $i = 1$, $W_{i-1} = \phi$, and so we only need check that $f[W_i] + f[u] \leq f[W_i] + f[u]$, which is clearly true. Suppose that the result holds for all $1 \leq i < k$. Consider any $S \subseteq W_{k-1}$. Let $j$ be the smallest integer such that $S \subseteq W_{j-1}$. If $j = k$, then $v_{k-1} \in S$ and $W_{k-1} \setminus S \subseteq W_{k-2}$. Therefore,

$$f[W_k \setminus S] + f[S \cup u]$$
$$\overset{(A)}{=} f[(W_{k-1} \setminus S) \cup v_k] + f[S \cup u]$$
$$\overset{(B)}{\geq} f(W_{k-1}) + f(v_k) - f(S) + f(S \cup u) - (k-2)\epsilon$$
$$\overset{(C)}{\geq} f(W_{k-1} \cup u) + f(v_k) - (k-1)\epsilon$$
$$\overset{(D)}{\geq} f(W_k) + f(u) - (k-1)\epsilon$$

Where (A) follows because $W_k \setminus S = (W_{k-1} \setminus S) \cup v_k$, (B) follows from the inductive hypothesis using $(v_k, W_{k-1} \setminus S)$ and $i = k - 1$ since $v_k \notin W_{k-1}$ and $W_{k-1} \setminus S \subseteq W_{k-2}$. (C) follows from approximate submodulartity by taking

$A = W_{k-1}$ and $B = S \cup u$. Then $A \cup B = W_{k-1} \cup u$ and $A \cap B = S$. To see (D), note that at step $k$, $v_k \in V \setminus W_{k-1}$ is selected so that $v_k = \arg\min_{u \in V \setminus W_{k-1}} (f[W_{k-1} \cup u] - f[u])$. Therefore, for $u \notin W_k$, we have $f[W_{k-1} \cup u] - f[u] \geq f[W_{k-1} \cup v_k] - f[v_k] = f[W_k] - f[v_k]$.

If $j < k$, then $v_{j-1} \in S$ and none of $v_j, \ldots, v_k$ are in $S$. We have

$$f[W_k \setminus S] + f[S + u]$$
$$= f[(W_{j-1} \setminus S) \cup (W_k \setminus W_{j-1})] + f[S \cup u]$$
$$\overset{(F)}{\geq} f[(W_{j-1} \setminus S) \cup (W_k \setminus W_{j-1})]$$
$$\quad + f[W_j] - f[W_j \setminus S] + f[u] - (j-2)\epsilon$$
$$\overset{(G)}{\geq} f[W_k] + f[u] - (j-1)\epsilon$$

Since $j < k$, we may use the inductive hypothesis with $(u, S)$ and $i = j$ to get (F). Then we use approximate submodularity by letting $A = (W_{j-1} \setminus S) \cup (W_k \setminus W_{j-1})$ and $B = W_j$. Then $A \cup B = W_k$ and $A \cap B = W_j \setminus S$. to get (G). Therefore, the claim holds by induction. □

**Theorem 14.** *Suppose that $f$ is approximately symmetric and submodular. Then the pendent pair $(t, u)$ returned Algorithm 3 satisfies $f[u] \leq \min_{\{U \subseteq V : u \in U, t \notin U\}} f[U] + \frac{n\epsilon}{2}$.*

*Proof.*

$$2f[v_n] - \epsilon \overset{(H)}{\leq} f[v_n] + f[W_{n-1}]$$
$$\overset{(I)}{\leq} f[W_{n-1} \setminus S] + f[S \cup v_n] + (n-2)\epsilon$$
$$\overset{(J)}{\leq} 2f[S \cup v_n] + (n-1)\epsilon$$

where $(H)$ follows from approximate submodularity, (I) follows from Lemma 13 with $i = n-1$ and $u = v_n$. (J) now follows by using approximate submodularity one again. □

Now, if $f$ is symmetric and submodular, and $\tilde{f}$ satisfies $\left| f(A) - \tilde{f}(A) \right| < \epsilon/4$, then $\tilde{f}$ is an approximately symmetric submodular function. An argument similar to proof of Theorem 3 in [10] now shows that Queyranne's algorithm will return a solution that is $n\epsilon$ close to optimal. Therefore, Lemma 4 follows.